# Finger Multimodal Feature Fusion and Recognition Based on Channel Spatial Attention

Jian Guo[1], Jiaxiang Tu[1], Hengyi Ren[1], Chong Han[1], Lijuan Sun[1]

[1]College of Computer, Nanjing University of Posts and Telecommunications, Nanjing 210003, China

## Abstract

Due to the instability and limitations of unimodal biometric systems, multimodal systems have attracted more and more attention from researchers. However, how to exploit the independent and complementary information between different modalities remains a key and challenging problem. In this paper, we propose a multimodal biometric fusion recognition algorithm based on fingerprints and finger veins (Fingerprint Finger Veins-Channel Spatial Attention Fusion Module, FPV-CSAFM). Specifically, for each pair of fingerprint and finger vein images, we first propose a simple and effective Convolutional Neural Network (CNN) to extract features. Then, we build a multimodal feature fusion module (Channel Spatial Attention Fusion Module, CSAFM) to fully fuse the complementary information between fingerprints and finger veins. Different from existing fusion strategies, our fusion method can dynamically adjust the fusion weights according to the importance of different modalities in channel and spatial dimensions, so as to better combine the information between different modalities and improve the overall recognition performance. To evaluate the performance of our method, we conduct a series of experiments on multiple public datasets. Experimental results show that the proposed FPV-CSAFM achieves excellent recognition performance on three multimodal datasets based on fingerprints and finger veins.

Keywords: Multimodal. Fingerprint. Finger vein. Channel Spatial Attention. Convolutional neural network.

## 1. Introduction

Biometric identification is the study of individual physiological and behavioral attributes in order to overcome security problems[1] . Common physiological and behavioral attributes include face[2], voice[3], iris[4], gait[5], fingerprint[6], veins[7], etc. Compared with traditional identification, such as keys, ID cards and passwords, it has versatility, stability, uniqueness and higher security.

Biometric recognition usually consists of four processes: (i) biometric acquisition; (ii) preprocessing; (iii) feature extraction; (iv) pattern recognition. In the feature extraction process, extracting suitable discriminative features from different single modalities can significantly improve the recognition performance. However, how to extract suitable discriminative biometrics for different single modalities remains a difficult problem. Existing biometric feature extraction methods can be roughly divided into two categories: traditional methods based[8-18] and deep learning based[19-25]. Traditional methods generally include subspace learning-based methods[10] and local feature-based approaches[12-17]. The subspace learning-based methods feature extraction approaches, such as principal component analysis (PCA)[10] and linear discriminant analysis (LDA)[26], consider the sub-space coefficients as discriminative features. However, the recognition efficiency and accuracy of the above methods are not very high. The local feature-based approaches, such as local binary patterns (LBP)[12], histogram of oriented gradient (HOG)[13], adaptive radius local binary pattern (ADLBP)[14], have been widely used for finger biometrics recognition. However, the above local feature-based approaches generally describe the relationship between image pixels and surrounding pixels, such as gradient, position information, etc., to capture the correlation of objects within a class and the irrelevance of objects between classes, but it is difficult to discover the hidden relationship between pixels. Therefore, only manually designed local feature description methods cannot effectively extract discriminative features of different single biological modalities. Compared with biometric feature extraction methods based on traditional methods, deep learning can automatically learn the relationship between pixels in images without human intervention, and the overall recognition performance is good. Therefore, researchers are increasingly focusing on deep learning based biometric feature extraction. Such as Pan *et al*.[23] designed a Multi-Scale Deep Representation Aggregation (MSDRA) model based on Deep Convolutional Neural Network (DCNN). MSDRA can also achieve good performance on small-scall finger vein datasets. Yang *et al*.[24] proposed Finger-Vein Recognition and AntiSpoofing Network (FVRAS-Net). FVRAS-Net adopts a multitask learning (MTL) approach to integrate the recognition task and the anti-spoofing task into a unified CNN model. Zhang *et al*.[25] proposed a lightweight convolutional neural network with a convolutional block attention module (CBAM). This network can reduce the amount of computation and consider the different importance of image pixels.

In practice, it has been found that unimodal biometric systems for authentication usually suffer from noisy sensor data, spoofing attacks, and lack of uniqueness[27, 28]. To overcome these shortcomings, researchers have begun to focus on multimodal biometric recognition techniques which fused different unimodal biometrics. Identification may be easier and more secure by exploiting the complementary information between different modal features. In this paper, this issue is also investigated and two biometric features, fingerprint and finger vein, are chosen. There are two reasons for choosing them: (1) Compared with other biological modalities, fingerprint and finger vein are easy to collect, with low costs of storage and computation. And they both come from the finger, so only one device is needed to collect these two biometrics at the same time. (2) Fingerprint is a surface feature and its recognition technology is relatively mature. Finger vein is an in internal feature which is not easy to be stolen. Hence, the combination of the two biometrics may enhance the accuracy and security of the identification system.

An ideal multimodal biometric fusion method should have such advantages as simple image preprocessing, efficient multimodal fusion, and good recognition performance. Currently, the fusion of multi-feature biometrics may be performed at three levels[29], namely pixel level[30], feature level[31], and score level[32]. At these different levels, researchers have proposed several multimodal biometric fusion methods[33-37]. However, most of these methods either do not extract the discriminative unimodal features sufficiently, or do not utilize the complementary information between different unimodal features fully, or are too complex or time-consuming. Therefore, finding the discriminative biometrics of different single biological modalities and combining the complementary information between different biometrics is still a key issue in biometric identification. To the above problems, a Channel Spatial attention based finger multimodal feature fusion method (FPV-CSAFM) is proposed, which integrates fingerprint and finger vein features.

The main contribution of this method is as follows.

• We design a dual-channel CNN to extract unimodal features concurrently and each unimodal features would not be influenced by others.

• We propose a multimodal feature fusion module (CSAFM). First, the different multimodal features use element-wise summation to complete the initial feature integration. Then, the multimodal feature obtains the corresponding channel attention fusion coefficient through the channel attention module, and inputs the coefficient into the spatial attention module to obtain the corresponding spatial attention fusion coefficient. The reason for this is to reduce the information reduction after feature vectors are re-corrected by channel dimension, amplifying the global interaction of channel and spatial dimension.

• By combining a dual-channel CNN and CSAFM, our multimodal biometric fusion recognition algorithm (FPV-CSAFM) is formed. Since the finger image information input into the CNN will be transformed into the corresponding feature vector, FPV-CSAFM can apply any two or more biological modalities by only adding additional CNN and modifying the vector number of the initial feature aggregation in CSAFM It has good scalability and robustness.

• FPV-CSAFM model performance was assessed across three fingerprints finger veins datasets.

The rest of this paper is organized as follows. In Section 2, some related studies are discussed. The proposed algorithm is presented in detail in Section 3. Experimental results and analysis are given in Section 4. Finally, conclusions are drawn in Section 5.

## 2. Related works

This section will introduce some multimodal fusion recognition methods, especially hand multimodal recognition methods. In [38],CNN has higher recognition accuracy. The attention mechanism has been shown to not only help CNNs improve recognition performance, but also tell CNNs where to focus, enhance important features and suppress unnecessary ones. Therefore, this section also introduces the attention mechanism.

### 2.1  Hand Multimodal Recognition

The core of the multimodal biometric fusion methods is feature extraction and fusion, and researchers have proposed some fusion methods. Zhong *et al.*[33] proposed a multimodal recognition algorithm (Palmprint Hybridized with Dorsal hand vein, PHD) based on palm print and dorsal hand vein (DHV) and two methods to improve the recognition rate of PHD. PHD uses deep hashing network (DHN) and biometric graph matching (BGM) to handle palm print and DHV, respectively. The first method to improve the recognition rate of PHD is to combine different features extracted from DHN and BGM. The second method uses DHN to process palm print and DHV, and implement PHD at pixel level, feature level and score level. However, the network structure used by it is relatively complex, resulting in a long recognition time. Yang *et al.*[34] proposed a fingerprint and finger-vein based cancelable multi-biometric system. The system uses three different feature-level fusion strategies to combine minutiae-based fingerprint feature sets and image-based finger vein feature sets. The system uses the enhanced partial discrete Fourier transform (EP-DFT) based non-invertible transformation to strengthen security. However, the EP-DFT based non-invertible transformation takes a long time, resulting in a slower overall recognition speed of the system. Daas *et al.*[35] proposed two multimodal architectures using the finger knuckle print (FKP) and the finger vein. Extract finger vein and FKP features using pretrained AlexNet, VGG16[39] and ResNet50[40], then fuse the two modalities at feature-level using concatenation or addition or at score-level using weighted product, weighted sum, or Bayesian rule Fusion of the two modalities. Although using a pre-trained classical network model as a feature extractor has a certain effect, a deeper network model means that the extracted features are more abstract and sparser, so part of the image texture information will be lost, resulting in a low recognition rate. Li *et al.*[36] proposed a convolutional neural network based on discriminative local coding (LC-CNN) to perform multimodal finger recognition by fusing fingerprint, finger vein and knuckle fingerprint features. LC-CNN uses a weighted local coding operator to enhance the discriminability of tri-modal finger images, and reconstructed it with a set of convolutional layers. Although CNN is used to implement the proposed local encoding operator more concisely, the operator cannot mine all the hidden relationships between image pixels, so the recognition performance is not particularly good. Yang *et al.*[37] proposed a novel method for human identification using feature-level fusion of finger veins and finger dorsal texture (FDT). The method uses α-trimmed Weber representation (α-TWR) to enhance the vessels underneath skin and line-like texture on skin. Cross section asymmetric coding (CSAC) was performed to extract features for each pixel, and local density weighted matching was developed to obtain a matching score between the two feature maps. In the fusion strategy of this method, FV and FDT are equivalent, and are not fused according to the image importance of different modalities. To be efficient and able to dynamically adjust the fusion weights according to the importance of different modalities during CNN training, we propose FPV-CSAFM.

## 2.2 Attention Mechanism

Attention mechanisms used to assist convolutional neural networks to focus on certain regions of an image and suppress less prominent pixels or channels have been extensively studied[41-47]. Hu *et al.*[41] proposed to use channel attention to perform feature recalibration, through which the channel attention mechanism can learn to use global information to selectively emphasize useful informative features and suppress less useful informative features. Liu *et al.*[42] proposed a novel normalization-based attention module (NAM), which suppresses less salient weights. NAM applies a weight sparsity penalty to the attention modules, making them more computational efficient while retaining similar performance. Hou *et al.*[43] proposed a novel attention mechanism for mobile networks, called coordinate attention. Coordinate attention factorizes channel attention into two 1D feature encoding processes that aggregate features along the two spatial directions, respectively. Woo *et al.*[44] proposed a simple and effective attention module for feed-forward convolutional neural networks, called Convolutional Block Attention Module (CBAM). Given an intermediate feature map, CBAM sequentially infers attention maps along two separate dimensions, channel and spatial, then the attention maps are multiplied to the input feature map for adaptive feature refinement. Park *et al.*[45] proposed a simple and effective attention module, named Bottleneck Attention Module (BAM). BAM infers an attention map along two separate pathways, channel and spatial. Liu *et al.*[46] proposed a global attention mechanism (GAM). GAM boosts the performance of deep neural networks by reducing information reduction and magnifying the global interactive representations. Dai *et al.*[47] proposed attentional feature fusion (AFF) and iterative attentional feature fusion (iAFF) to better fuse features of inconsistent semantics and scales. Experimental results show that both AFF and iAFF can improve the performance of existing network models.

However, the above attention mechanisms either only consider the attention of a certain dimension (channel dimension or spatial dimension), or do not consider the interaction relationship between channel and spatial attention, and do not consider the information interaction between different dimensions, so some cross-dimensional information may be lost. To overcome the above shortcomings and apply the attention mechanism to multimodal fusion recognition, we propose CSAFM.

## 3. The proposed method

In this section, we first introduce the proposed FPV-CSAFM. Based on this, we detail the CNN unimodal network used by FPV-CSAFM, which is specialized for extracting fingerprint or finger vein features with discriminative features. Finally, in Section 3.3, we introduce the Channel-Spatial Attention Fusion Module (CSAFM) used in FPV-CSAFM for feature extraction and feature fusion of multimodal images of fingers.

### 3.1 FPV-CSAFM

Given fingerprint and finger vein images, we first use CNN unimodal network to extract the discriminative features of fingerprint and finger vein respectively, and then input the two features into CSAFM. Finally, take the output of CSAFM as the input of Softmax, and output the probability of each category to which the test sample belongs. Fig.1 illustrates the overall architecture of our network, and a detailed explanation of the CNN unimodal network and CSAFM can be found in Sections 3.2 and 3.3.

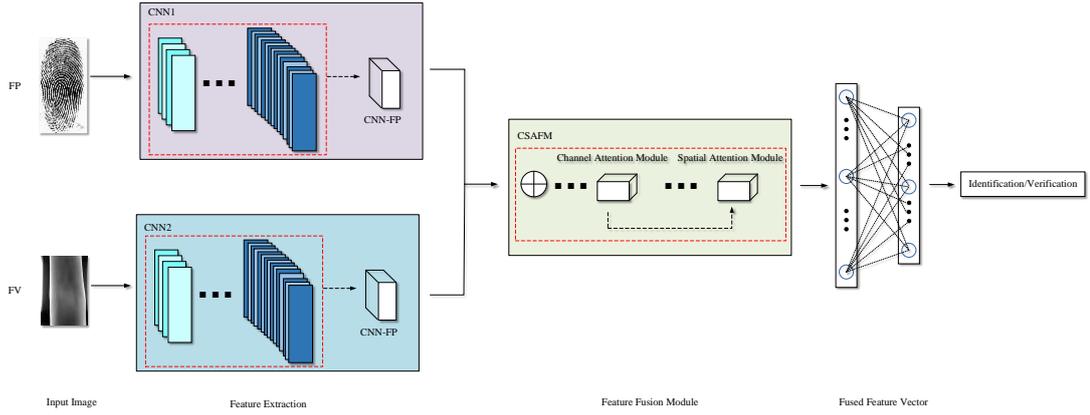

Fig.1. The basic idea of the FPV-CSAFM

## 3.2 CNN unimodal network

Today's popular CNN network models such as VGG[39], ResNet[40], ResNeXt[48] often require a large amount of training data to achieve optimal results. With deepening the model layers, the features extracted by CNN are becoming more and more abstract, and the degree of sparseness is also increasing. Although the model can fit more complex feature input, it also means that the model will lose a lot of detailed texture information of the original image. In the finger recognition task, the data set is usually relatively small. If we simply apply the above popular CNN deep network model, the origin texture information of the image will be easily lost, and even overfitting will occur, resulting in a low overall recognition rate. However, whether it is a fingerprint image or a finger vein image, the features contained in the image are relatively small, so shallow networks may be more suitable for feature extraction. So, we propose a novel CNN unimodal network for feature extraction from fingerprint or finger vein images. The proposed CNN unimodal network architecture consists of 5 convolutional layers, 5 Rectified Linear Unit (ReLU) layers, 5 Batch Normalization (BN) layers, 5 max-pooling layers, 1 flatten layer and 1 fully-connected layer. Details of the parameters are listed in Table 1 and the architecture of CNN unimodal network is shown in Fig.2.

Table 1
The parameters of CNN unimodal network

| Layer | Filter Size | Stride |
| --- | --- | --- |
| Conv1 | 7×7×64 | 2 |
| BN1 | 64 | - |
| ReLU1 | - | - |
| Max-pool1 | 3×3 | 2 |
| Conv2 | 3×3×128 | 1 |
| BN2 | 128 | - |
| ReLU2 | - | - |
| Max-pool2 | 3×3 | 2 |
| Conv3 | 3×3×256 | 1 |
| BN3 | 256 | - |
| ReLU3 | - | - |
| Max-pool3 | 3×3 | 2 |
| Conv4 | 3×3×512 | 1 |
| BN4 | 512 | - |
| ReLU4 | - | - |
| Max-pool4 | 3×3 | 2 |
| Conv5 | 3×3×512 | 1 |
| BN5 | 512 | - |
| ReLU5 | - | - |
| Max-pool5 | 3×3 | 2 |
| Flatten | - | - |
| FC | 1×1×classes | 1 |

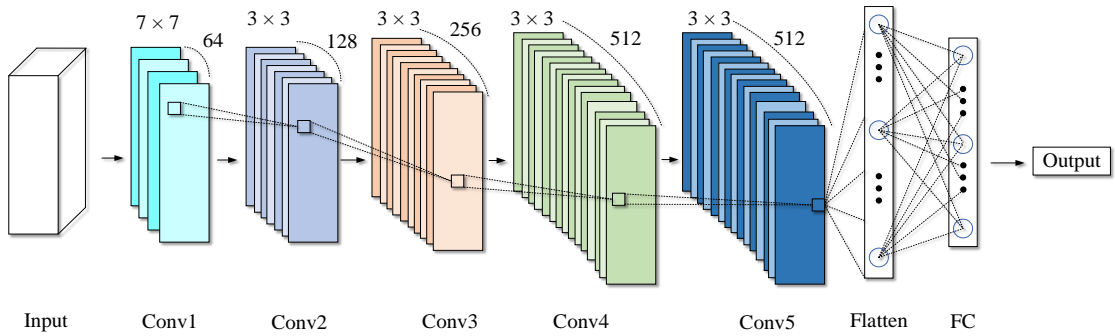

Fig.2. The architecture of CNN unimodal network

## 3.3 CSAFM

In this section, we describe the proposed Channel-Spatial Attention Fusion Module (CSAFM) in detail. Driven by various attention mechanisms[41-47], the recognition performance of CNN network continues to improve. However, the above attention mechanism is either only applied to a certain dimension such as channel dimension, spatial dimension, or does not consider the information interaction between different dimensions, which splits the interaction between dimensions and loses cross-dimensional information. Our research goal is to use the attention mechanism to fully mine the complementary information between different modalities, amplify the interaction between the channel and the spatial dimension, so as to better combine the information between different modalities and improve the recognition performance Therefore, we propose CSAFM. The overall architecture of CSAFM is shown in Fig.3, including the channel attention module shown in Fig.4(a) and the spatial attention module shown in Fig.4(b).

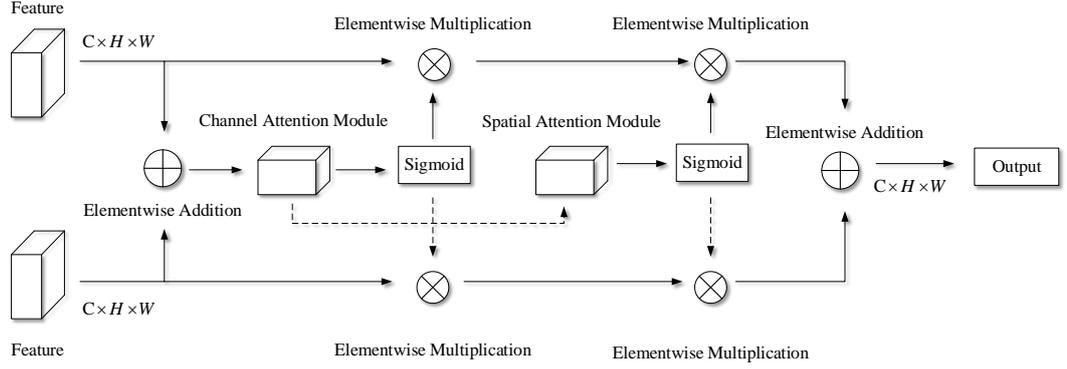

Fig.3. The overview of CSAFM

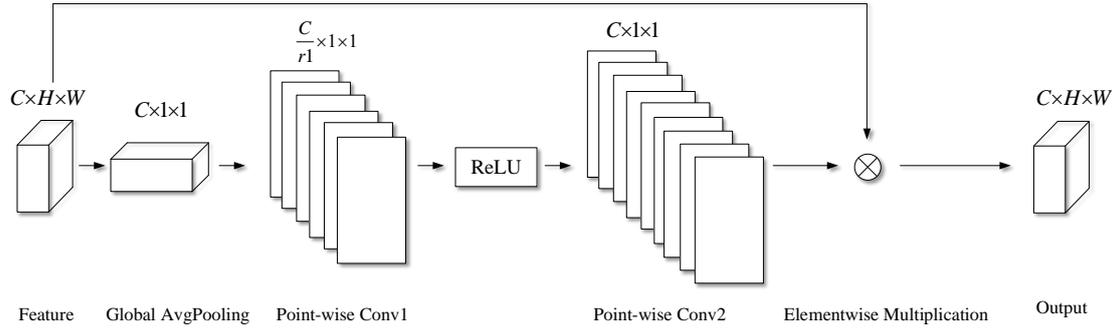

(a)

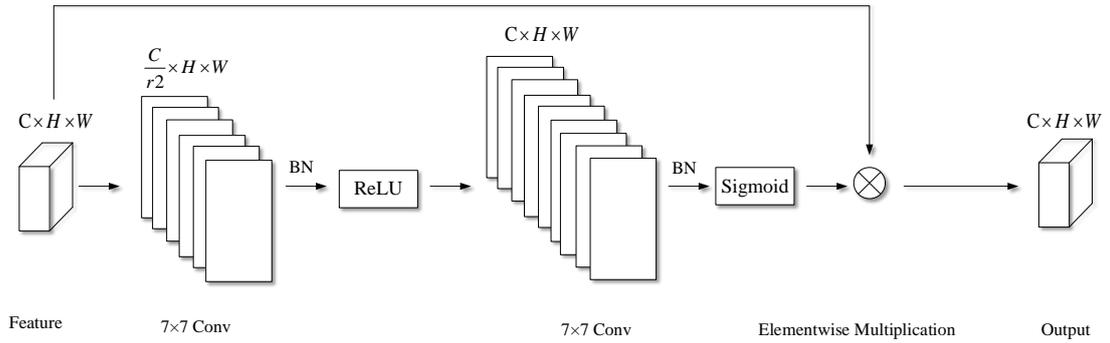

(b)

Fig.4. The channel and spatial attention module included in CSAFM (a) Channel attention module (b) Spatial attention module

In the channel attention module, to efficiently compute channel attention, we first compress the spatial dimension of the input feature map by global average pooling (GAP). Because point-wise convolution (PWConv) can achieve cross-channel interaction and information integration, reducing the number of convolution channels while reducing information loss. Therefore, we apply PWConv for channel rearrangement according to the importance of different channels in the features. We adopt the same reduction ratio *r1* as SE Block[41] . Specifically, given an intermediate feature $X \in \mathbb{R}^{C \times H \times W}$ with *C* channels and feature maps of size $H \times W$. The output of the channel attention module $C(X) \in \mathbb{R}^{C \times H \times W}$ can be computed as

$$C(X) = X \otimes PWConv2(\delta(PWConv1(g(X)))), \tag{1}$$

where $g(X) \in \mathbb{R}^C$ is GAP, defined as:

$$g(X) = \frac{1}{H \times W} \sum_{i=1}^{H} \sum_{j=1}^{W} X_{[:,i,j]}, \tag{2}$$

The kernel sizes of PWConv1 and PWConv2 are $\frac{C}{r1} \times 1 \times 1$ and $C \times 1 \times 1$, respectively. $\delta$ denotes ReLU, $\otimes$ denotes the element-wise multiplication。

We take the output of the channel attention module as the input of the spatial attention module. In the spatial attention module, we use two 7×7 convolutions to expand the receptive field and perform spatial information fusion. We adopt the same reduction ratio $r2$ as BAM[45]. Specifically, given an intermediate feature $X \in \mathbb{R}^{C \times H \times W}$ with $C$ channels and feature maps of size $H \times W$. The output of the spatial module $S(X) \in \mathbb{R}^{C \times H \times W}$ can be computed as

$$S(X) = X \otimes \sigma(\beta(Conv2(\delta(\beta(Conv1(X))))))), \tag{3}$$

The kernel sizes of Conv1 and Conv2 are $\frac{C}{r2} \times 7 \times 7$ and $C \times 7 \times 7$, respectively. $\beta$ denotes BN, $\sigma$ is the Sigmoid function.

Based on the channel attention module and the spatial attention module, given the fingerprint and finger vein feature vectors $F_{fp} \in \mathbb{R}^{C \times H \times W}$ and $F_{fv} \in \mathbb{R}^{C \times H \times W}$ extracted by the CNN unimodal network proposed in Section 3.1, CSAFM first standardizes the two feature vector dimensions. In order to retain as much feature information as possible, choose the largest overlapping area in the fingerprint and finger vein feature vectors is used as a standardized size, and the standardized definition is:

$$H = \min(H_{fp}, H_{fv}), \tag{4}$$

$$W = \min(W_{fp}, W_{fv}), \tag{5}$$

Then, we choose element-wise summation as the Initial Feature Integration (IFI) method. The formula for calculating IFI is:

$$IFI = F_{fp} \oplus F_{fv}, \tag{6}$$

$\oplus$ denotes the element-wise summation.

Put the features after the initial feature integration into the channel attention module to get the intermediate state $F_c$. $F_c$ is defined as:

$$F_c = C(IFI) \otimes IFI, \tag{7}$$

where $C$ corresponds to formula (1);

Substitute $F_c$ into sigmoid() to get the fusion coefficient $F_{c\_final}$, which is a real number between 0 and 1.

Put $F_c$ into the spatial attention module to get another intermediate state $F_s$. $F_s$ is defined as:

$$F_s = S(F_c) \otimes F_c, \tag{8}$$

where $S$ corresponds to formula (3);

Substitute $F_s$ into sigmoid() to get the fusion coefficient $F_{s\_final}$, which is a real number between 0 and 1.

Finally, after the CSAFM multimodal fusion module, the output $Z$ is defined as:

$$Z = F_{fp} \otimes F_{c\_final} \otimes F_{s\_final} \oplus F_{fv} \otimes (1 - F_{c\_final}) \otimes (1 - F_{s\_final}), \tag{9}$$

It should be noted that the fusion weights $F_{c\_final}$ and $F_{s\_final}$ consists of real numbers between 0 and 1, so are the $(1 - F_{c\_final})$ and $(1 - F_{s\_final})$, which enable the network to conduct a soft selection or weighted averaging between $F_{fp}$ and $F_{fv}$.

## 4. Experiments and results

In Section 4.1, we first introduce the fingerprint finger vein multimodal database used for experiments. Then, we discuss the parameter setting of the proposed method in Section 4.2. To verify the effectiveness of the CNN unimodal network proposed in Section 3.1, the proposed CNN unimodal network is compared with some existing unimodal recognition methods in Section 4.3. In Section 4.4, we conduct a series of ablation experiments in order to verify the effect of the channel and spatial attention modules in CSAFM. In Section 4.5, to explore the effectiveness of CSAFM, we first conduct a series of experiments using different combinations of single-modal extraction networks and CSAFM. Then, we compare our proposed FPV-CSAFM with state-of-the-art finger fusion recognition approaches.

### 4.1 Databases

NUPT-FPV[49] : In this dataset, the researchers simultaneously collected 16800 fingerprint images and 16800 finger vein images from 140 volunteers(two sessions). The 140 volunteers ranged in age from 16 to 39, with 108 and 32 men and women, respectively. We used 8400 fingerprint images and 8400 finger vein images of 140 volunteers from session 1. We use samples of the ROI version provided by the dataset with a 200 × 400 and 300 × 450, respectively.

MMCBNU_6000[50, 51] : In this dataset, researchers collected 6000 finger vein images of 100 volunteers. The 100 volunteers came from 20 countries/regions. Images were taken of each volunteer's index finger, middle finger, and ring finger on both two hands, generating a total of 600 finger categories and 10 samples for each finger. We use the original and ROI version samples provided by the dataset with resolutions: 640 × 480 and 128 × 60, respectively.

By combining the top 600 fingerprint images in NUPT-FPV and MMCBNU_6000, we construct two multimodal finger databases: MMCBNU-Origin-FPV and MMCBNU-ROI-FPV, respectively. More details of the databases can be found in Table 2. Fig.5 shows some samples of the original images and ROI images selected from the three different databases.

Table 2

Details of NUPT-FPV, MMCBNU-Origin-FPV and MMCBNU-ROI-FPV datasets

| Database | Subject number | Image per subject | Image size | Total image num |
|---|---|---|---|---|
| Data1(NUPT-FPV-Origin-FP) | 840 | 10 | 300 × 400 | 8400 |
| Data2(NUPT-FPV-Origin-FV) | 840 | 10 | 1920 × 1080 | 8400 |
| Data3(NUPT-FPV-ROI-FP) | 840 | 10 | 200 × 400 | 8400 |
| Data4(NUPT-FPV-ROI-FV) | 840 | 10 | 300 × 450 | 8400 |
| Data5(MMCBNU-Origin-FV) | 600 | 10 | 640 × 480 | 6000 |
| Data6(MMCBNU-ROI-FV) | 600 | 10 | 128 × 60 | 6000 |

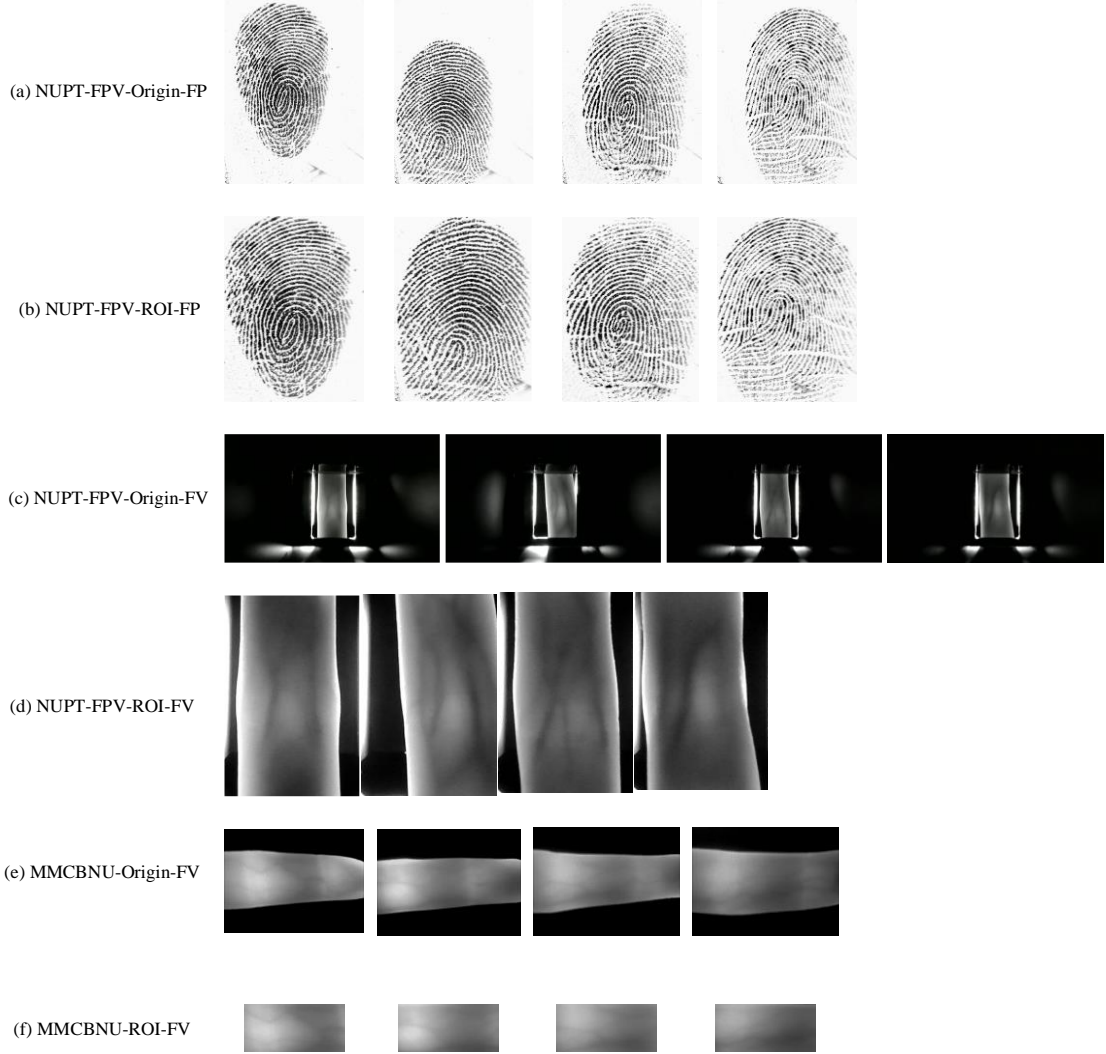

Fig.5. The multimodal finger images. (a) Origin fingerprint images of NUPT-FPV dataset. (b) Fingerprint ROI images of NUPT-FPV dataset. (c) Origin finger vein images from NUPT-FPV dataset. (d) Finger vein ROI images of NUPT-FPV dataset；(e) Origin images of finger veins from the MMCBNU_6000 dataset. (f) Finger vein ROI images of MMCBNU_6000 dataset

### 4.2 Parameter settings

In the proposed FPV-CSAFM model, adaptive moment estimation (Adam) was used to calculated the loss for optimization. The learning rate of FPV-CSAFM was set at 0.0001 with a batch size of 32, and the number of epochs was defined as 100 for the following experiments. For training purposes, 30% of multimodal finger images are selected as training samples, 40% as validation samples, and the remaining 30% as test samples. More details of the training, validation and testing datasets are shown in Table 3.

Table 3
Training, validating and testing multimodal databases

| Dataset | Percentage | Images | Categories |
|---|---|---|---|
| Training(NUPT-FPV) | 30% | $3 \times 840 \times 2$ | 840 |
| Validating(NUPT-FPV) | 40% | 4 × 840 × 2 | 840 |
| Testing(NUPT-FPV） | 30% | 3 × 840 × 2 | 840 |
| Training(MMCBNU-Origin-FPV) | 30% | 3 × 600 × 2 | 600 |
| Validating(MMCBNU-Origin-FPV) | 40% | 4 × 600 × 2 | 600 |
| Testing(MMCBNU-Origin-FPV) | 30% | 3 × 600 × 2 | 600 |
| Training(MMCBNU-ROI-FPV) | 30% | 3 × 600 × 2 | 600 |
| Validating(MMCBNU-ROI-FPV) | 40% | 4 × 600 × 2 | 600 |
| Testing(MMCBNU-ROI-FPV) | 30% | 3 × 600 × 2 | 600 |

In this paper, we used CIR (correct identification rate) to evaluate the performance of the method. CIR is calculated by the following formula:

$$CIR = \frac{correct\ predicted\ samples}{total\ samples} *100\%, \quad (10)$$

The CNN was trained on a single NVIDIA 2080Ti using the PyTorch deep learning framework, and the rest of the experiments were implemented on MATLAB 2017a with CPU; System configuration was: an Intel(R) Core(TM) i7-9700K CPU 3.60GHz, NVIDIA RTX 2080Ti GPU and 64 Bit Windows10 operating system.

### 4.3  Unimodal identification

To verify the effectiveness of the proposed CNN unimodal network (proposed in Section 3.1), we use the CNN unimodal network to extract fingerprint and finger vein features on the multimodal dataset as described in Section 4.1.2, and use Softmax to classify the extracted features. Combining the above network model with 1. Using traditional feature descriptors Local Binary Pattern (LBP)[12], Histogram of Oriented Gradient (HOG)[13] and Adaptive Radius Local Binary Pattern (ADLBP)[14] as a feature extractor, use K-Nearest Neighbor Algorithm (KNN) as a classifier. 2. Different classic CNN network models such as VGG16[39], ResNet18[40], ResNet50[40], ResNeXt[48] are compared. Using the setup explained in Section 4.2, the experimental results are shown in Table 4, Fig.6 and Fig.7.

Table 4
The performance of unimodal recognition (the fingerprint datasets used by MMCBNU-Origin-FPV and MMCBNU-ROI-FPV are the same, so the corresponding two columns have the same content)

| Modal | Method | CIR | | |
|---|---|---|---|---|
| | | NUPT − FPV | MMCBNU − Origin − FPV | MMCBNU − ROI − FP |
| FV | LBP+KNN | 77.96% | 97.00% | 97.35% |
| FV | HOG+KNN | 67.39% | 90.28% | 97.47% |
| FV | ADLBP+KNN | 85.76% | 96.11% | 97.76% |
| FV | VGG16 | 79.31% | 93.21% | 88.60% |
| FV | ResNet18 | 85.71% | 94.48% | 97.52% |
| FV | ResNet50 | 81.80% | 93.90% | 94.38% |
| FV | ResNeXt50 | 82.02% | 93.98% | 94.79% |
| FV | **Proposed method** | **86.96%** | **98.45%** | **98.74%** |
| FP | LBP+KNN | 82.43% | 80.85% | 80.85% |
| FP | HOG+KNN | 76.58% | 75.38% | 75.38% |
| FP | ADLBP+KNN | 79.01% | 76.40% | 76.40% |
| FP | VGG16 | 68.44% | 69.45% | 69.45% |
| FP | ResNet18 | 86.02% | 81.67% | 81.67% |
| FP | ResNet50 | 83.83% | 78.88% | 78.88% |
| FP | ResNeXt50 | 85.10% | 84.33% | 84.33% |
| FP | **Proposed method** | **91.58%** | **91.67%** | **91.67%** |

Table 4 summarizes the performance results of the proposed CNN unimodal network with some existing traditional algorithms or classical CNN network models. Our method achieves the highest accuracy in both fingerprint datasets and finger vein datasets. Experimental results show that our CNN single-modal network can mine more discriminative biometrics than traditional algorithms or classical network models. Compared with ResNet18, ResNet50 deepens the model layers and increases the number of model parameters, but the recognition effect decreases. This is because as the number of network layers deepens, the feature vector extracted by CNN will become more and more abstract, the dimension will become smaller and smaller, and the degree of sparseness will become higher and higher. Therefore, the deeper the CNN network model, the easier it is to lose the detailed texture information of the original image itself. For fingerprints and finger vein images, the detailed texture information is an important key factor to distinguish the same or different type. Therefore, for the unimodal finger image itself, it is not good to use a network model with deep layers and large number of parameters. This also illustrates the necessity of designing a CNN unimodal network with shallow network layers, small network model parameters, and suitable for fingerprints or finger veins.

The training process of the unimodal recognition network for fingerprints and finger veins is shown in Fig.6 and Fig.7. With the increase of the number of network training epoch, the CIR of both finger veins and fingerprints slowly fluctuated up.

In the unimodal recognition results of the NUPT-FPV dataset, the recognition results of most fingerprints are higher than those of finger veins. This is because the fingerprint image contains more texture features, and the image of the finger vein is greatly interfered by the external environment due to the limitation of the acquisition equipment and conditions, resulting in a relatively poor recognition performance of the finger vein images. However, from the perspective of security, fingerprints exist on the surface of human skin and are easily disturbed by the external environment of watch surface loss, dryness or excessive humidity. Finger veins are the internal characteristics of the human body, and their security and stability are higher than fingerprints. Therefore, it is meaningful to explore a multi-modal fusion recognition method for fingers that combines fingerprints and finger veins.

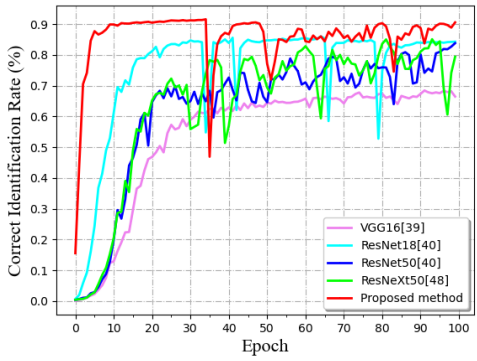
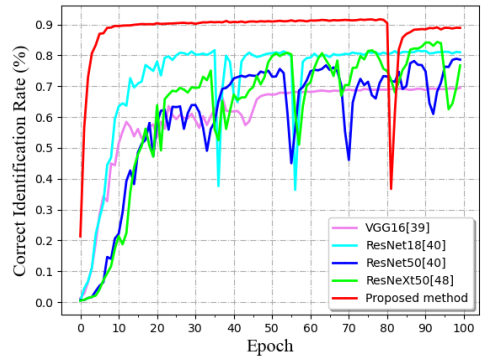

(a) NUPT-FPV  (b) MMCBNU-Origin-FPV

Fig.6. The training process of the two fingerprint datasets of experiments. The x-axis is the training epoch, and the y-axis is the CIR. Among them, (a) Fingerprint CIR of NUPT-FPV. (b) Fingerprint CIR of MMCBNU-Origin-FPV

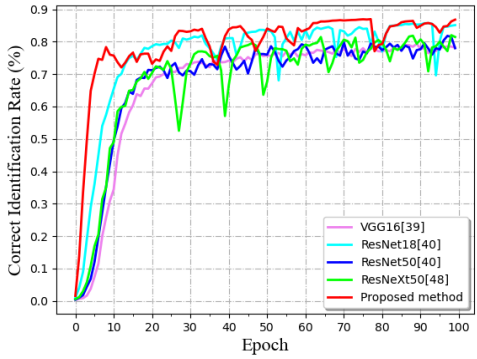
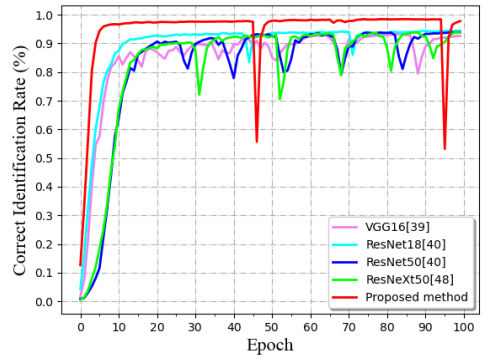

(a) NUPT-FPV  (b) MMCBNU-Origin-FPV

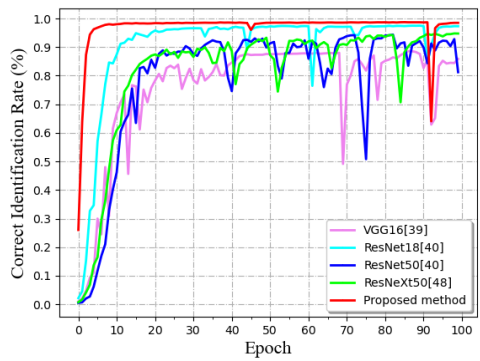

(c) MMCBNU-ROI-FPV

Fig.7. The training process of the three finger veins datasets of experiments. The x-axis is the training epoch, and the y-axis is the CIR. Among them, (a) Finger Veins CIR of NUPT-FPV (b) Finger Veins CIR of MMCBNU-Origin-FPV. (c) Finger Veins CIR of MMCBNU-ROI-FPV

## 4.4 Ablation studies

In order to study the compositional order and necessity of the channel and spatial attention modules in CSAFM, in this section, we construct four different ablation studies modules: independent channel attention module, independent spatial attention module, using channel and spatial attention modules in parallel, and using spatial attention module and channel attention module sequentially. The CSAFM proposed in this paper is referred to here as: sequentially using channel attention module and spatial attention module.

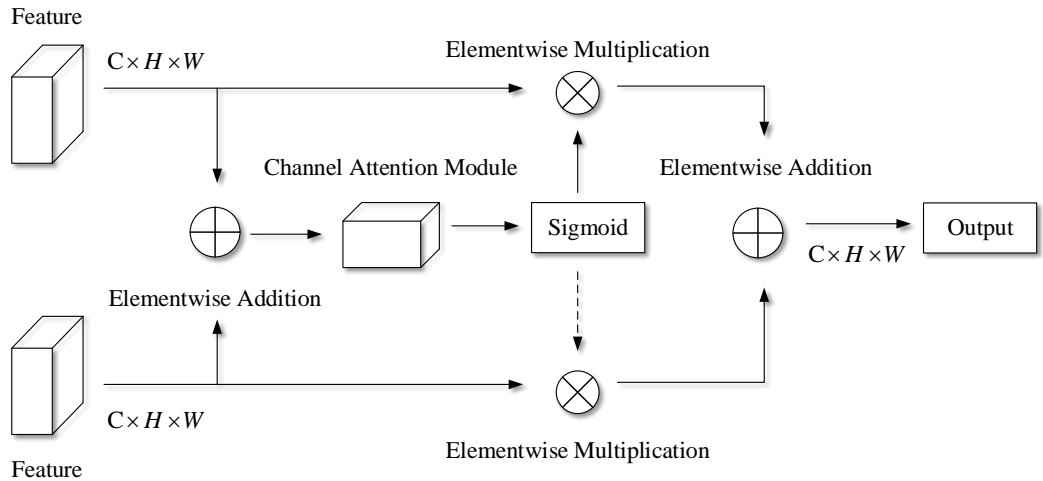

(a)

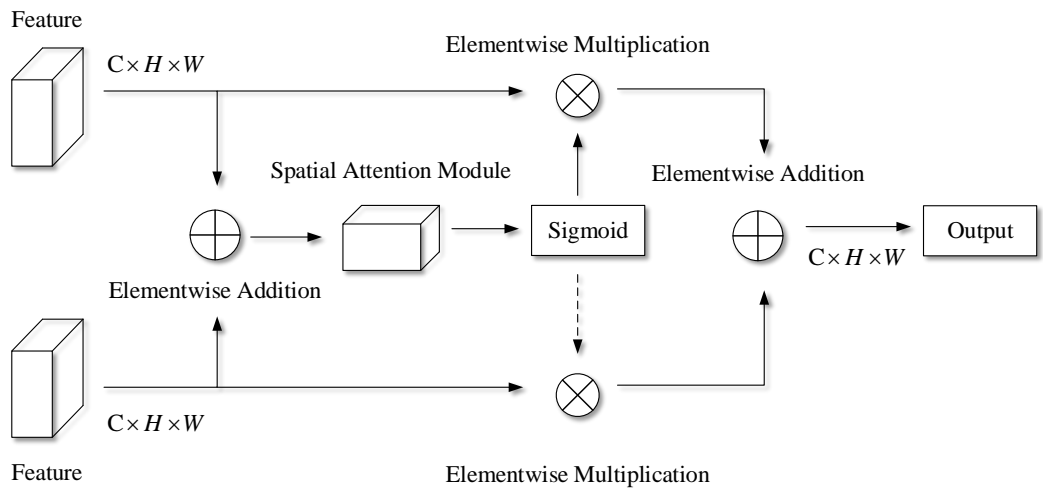

(b)

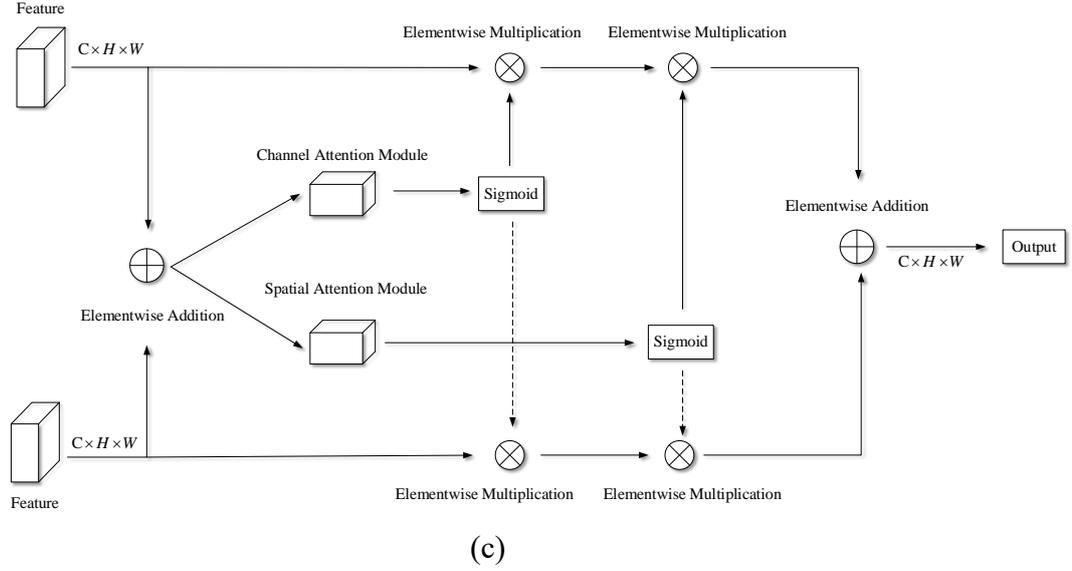

(c)

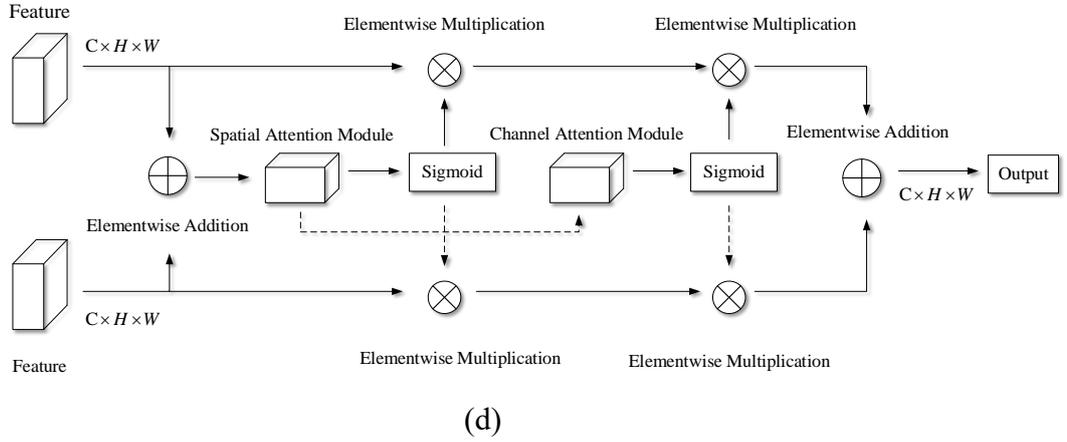

(d)

Fig.8. Ablation study modules with different combinations (a) independent channel attention module (b) independent spatial attention module (c) using channel and spatial attention modules in parallel (d) using spatial attention module and channel attention module sequentially

Table 5

Comparison of different modules combinations

| Module combination | CIR | | |
|---|---|---|---|
| | NUPT − FPV | MMCBNU − Origin − FPV | MMCBNU − ROI − FPV |
| independent channel attention module | 94.37% | 99.38% | 99.29% |
| independent spatial attention module | 92.74% | 99.33% | 98.88% |
| using channel and spatial attention modules in parallel | 93.69% | 99.50% | 98.98% |
| using spatial and channel attention module sequentially | 92.13% | 99.33% | 98.88% |
| Propose method | **95.41%** | **99.60%** | **99.76%** |

Table 5 shows the CIR of different combinations of the above four different ablation modules and our proposed CSAFM on the NUPT-FPV, MMCBNU-Origin-

FPV and MMCBNU-ROI-FPV datasets. From the results, we can find that no matter how many attention modules are used in the ablation module, the performance of preferentially using the channel attention module is better than that of preferentially using the spatial attention module. In fact, this is also in line with some aspects of human visual cognition, which is to focus on the "whole" (channel attention module) and then the "local" (spatial attention module). While the CIR of our proposed CSAFM outperforms the other four ablation modules, CSAFM should always be preferred for finger multimodal feature fusion.

### 4.5 Multimodal identification

#### 4.5.1 Effectiveness of CSAFM with Different Networks

In order to further verify the effectiveness of our proposed multimodal feature fusion module CSAFM, we choose different unimodal feature extraction networks such as ResNet18, ResNet50 and different combinations of multimodal feature fusion methods. The CIR are shown in Table 6, Fig.9 and Fig.10. From the results, we can see that although a different network model is used to replace our proposed CNN unimodal network, according to the comparison of different fusion strategies, CSAFM is the best among all fusion strategies. This indicates that CSAFM not only has excellent performance but also has good generality.

Table 6
CIR of different network models

| Network | Fusion method | CIR | | |
|---|---|---|---|---|
| | | NUPT-FPV | MMCBNU-Origin-FPV | MMCBNU-ROI-FPV |
| ResNet18 | Daas *et.al.*[35] | 90.41% | 97.14% | 97.48% |
| ResNet18 | Dai *et.al.*[47] | 90.71% | 97.62% | 98.29% |
| ResNet18 | Proposed method | 92.52% | 98.07% | 98.62% |
| ResNet50 | Daas *et.al.* [35] | 83.15% | 92.67% | 91.95% |
| ResNet50 | Dai *et.al.* [47] | 87.18% | 95.17% | 95.17% |
| ResNet50 | Proposed method | 88.95% | 95.24% | 95.52% |

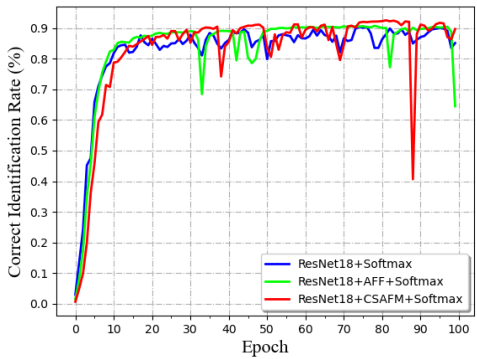

（a） NUPT-FPV

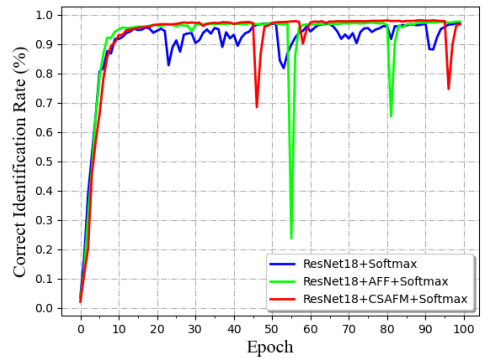

(b) MMCBNU-Origin-FPV

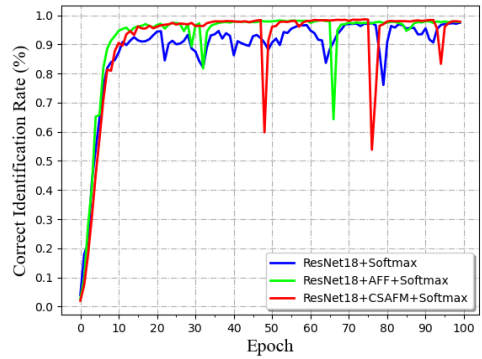

(c) MMCBNU-ROI-FPV

Fig.9. The training process of ResNet18 multimodal different fusion recognition methods. The x-axis is the training epoch, and the y-axis is the CIR. Among them, (a) CIR of NUPT-FPV (b) CIR of MMCBNU-Origin-FPV (c) CIR of MMCBNU-ROI-FPV

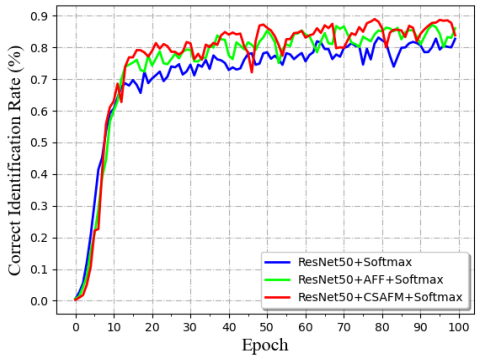

(a) NUPT-FPV

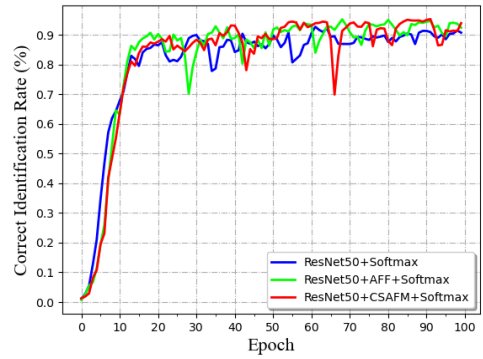

(b) MMCBNU-Origin-FPV

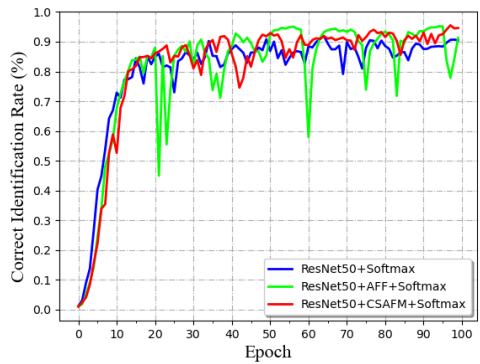

(c) MMCBNU-ROI-FPV

Fig.10. The training process of ResNet50 multimodal different fusion recognition methods. The x-axis is the training epoch, and the y-axis is the CIR. Among them, (a) CIR of NUPT-FPV (b) CIR of MMCBNU-Origin-FPV (c) CIR of MMCBNU-ROI-FPV

### 4.5.2 Comparison with traditional algorithms

We choose traditional feature descriptors LBP[12] , HOG[13] , ADLBP[14] as feature descriptors, and then use two different feature fusion methods 1. Serial (two feature vectors are summed element-wise to form a fused feature vector. The fused feature vector has the same dimension as the original feature vector.) 2. Parallel (two features are spliced together to form a fused feature vector, and the dimension of the fused feature vector is twice that of the original feature vector). Finally use KNN as the classifier to get the final result. The CIR of the unimodal, multimodal serial, parallel and our proposed algorithm are shown in Table 7.

Table 7
CIR for different modalities, different algorithms and combinations of classifiers

| | | *CIR* | | |
|---|---|---|---|---|
| Modal | Method | NUPT − FPV | MMCBNU − Origin − FPV | MMCBNU − ROI − FPV |
| FV | LBP+KNN | 77.96% | 97.00% | 97.35% |
| FP | LBP+KNN | 82.43% | 80.85% | 80.85% |
| FV+FP(Serial) | LBP+KNN | 87.94% | 97.61% | 99.00% |
| FV+FP(Parallel) | LBP+KNN | 87.99% | 97.73% | 99.09% |
| FV | HOG+KNN | 67.39% | 90.28% | 97.47% |
| FP | HOG+KNN | 76.58% | 75.38% | 75.38% |
| FV+FP(Serial) | HOG+KNN | 82.11% | 94.50% | 98.28% |
| FV+FP(Parallel) | HOG+KNN | 82.39% | 94.97% | 98.95% |
| FV | ADLBP+KNN | 85.76% | 96.11% | 97.76% |
| FP | ADLBP+KNN | 79.01% | 76.40% | 76.40% |
| FV+FP(Serial) | ADLBP+KNN | 89.42% | 97.54% | 98.40% |
| FV+FP(Parallel) | ADLBP+KNN | 89.91% | 97.92% | 98.47% |
| FV | Proposed method | **86.96%** | **98.45%** | **98.74%** |
| FP | Proposed method | **91.58%** | **91.67%** | **91.67%** |
| FV+FP | Proposed method | **95.41%** | **99.60%** | **99.76%** |

From the results shown in Table 7, we can see different CIRs for different modalities, different algorithms and combinations of classifiers. Compared with unimodal recognition, whether LBP+KNN, HOG+KNN, ADLBP+KNN or the algorithm proposed in this paper have achieved better recognition results in multimodal recognition. This indicates that multimodal recognition can make full use of the complementary information between different modalities to improve the overall recognition performance, and it is meaningful to explore multimodal finger recognition work. Furthermore, the CIRs on multiple multimodal datasets show that the proposed method achieves the best recognition performance. This shows that our proposed CNN single-modality network can not only mine more discriminative biological features than traditional algorithms, but also show that our multi-modal fusion module CSAFM can better fuse the feature vectors of different modalities, using Complementary information between different modalities, thereby improving the overall recognition rate of the system.

### 4.5.3 Comparison with the state-of-the-art

In order to verify the advancedness of the proposed algorithm in this paper, we compare the proposed FPV-CSAMF with several other state-of-the-art biological multimodal fusion recognition algorithms[35, 36, 47, 49, 52-55]. Table 8 and Fig.11 shows the training process of different biological multimodal fusion recognition algorithms. On NUPT-FPV, MMCBNU-Origin-FPV, MMCBNU-ROI-FPV fingerprint finger vein multimodal datasets, the proposed FPV-CSAFM is superior in CIR compared with several other multimodal fusion recognition algorithms. In the comparison experiments, we do not directly replicate the results from the references, and all methods are replicated in our experimental environment. The same size dataset and hardware environment can ensure the fairness of different algorithms when comparing the results. Finally, the proposed FPV-CSAFM has a CIR of 95.41%, 99.60%, 99.76% on NUPT-FPV, MMCBNU-Origin-FPV and MMCBNU-ROI-FPV datasets.

Table 8
CIR of multimodal recognition

| Method | CIR | | |
|---|---|---|---|
| | NUPT − FPV | MMCBNU − Origin − FPV | MMCBNU − ROI − FPV |
| Dass *et.al.*[35] | 86.82% | 93.02% | 91.88% |
| Li *et al.*[36] | 83.69% | 93.98% | 94.64% |
| Dai *et.al.*[47] | 85.39% | 96.02% | 94.62% |
| Ren *et.al.*[49] | 75.77% | 86.62% | 80.05% |
| El *et.al.*[52] | 70.14% | 95.69% | 93.95% |
| Wang *et.al.*[53] | 80.41% | 93.36% | 94.07% |
| Soleymani *et.al.*[54] | 82.55% | 91.76% | 93.45% |
| Alay *et.al.*[55] | 80.97% | 91.26% | 91.02% |
| **Proposed method** | **95.41%** | **99.60%** | **99.76%** |

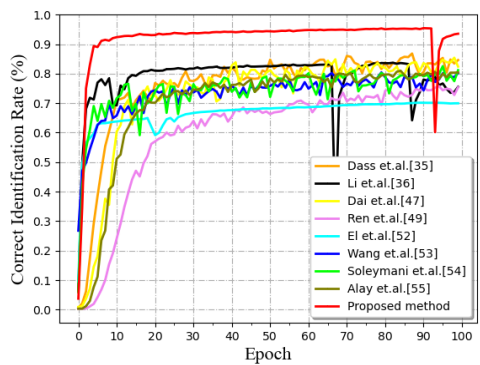

(a) NUPT-FPV

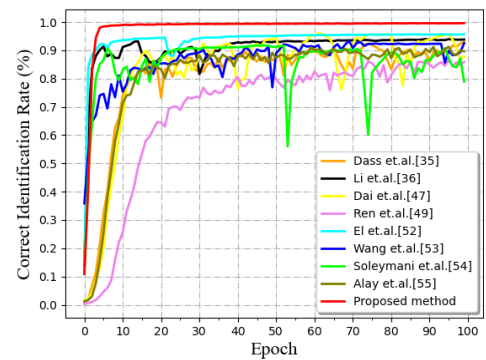

(b) MMCBNU-Origin-FPV

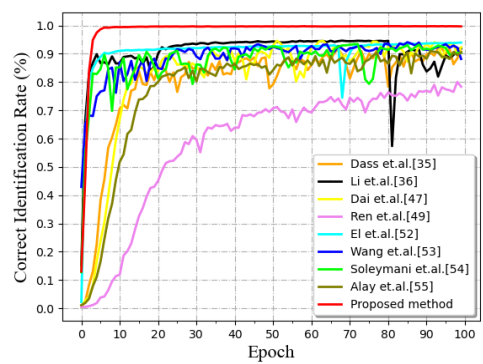

(c) MMCBNU-ROI-FPV

Fig.11. The training process of different multimodal recognition methods. The x-axis is the training epoch, and the y-axis is the CIR. Among them, (a) CIR of NUPT-FPV (b) CIR of MMCBNU-Origin-FPV (c) CIR of MMCBNU-ROI-FPV

## 5. Conclusion

This paper proposes a multimodal finger recognition method based on the channel-spatial attention mechanism. We first extracted and normalized fingerprint and finger vein features in parallel through the self-designed CNN unimodal network, and then fused the extracted fingerprint and finger vein features through the multimodal fusion module CSAFM, and finally obtained the classification results through Softmax. The experimental results on the fingerprint finger vein datasets NUPT-FPV, MMCBNU-Origin-FPV and MMCBNU-ROI-FPV show that the proposed method has better recognition results than some existing multimodal finger fusion recognition methods. In addition, the system proposed in this paper is highly scalable. Since the finger image information input into CNN will be converted into the corresponding feature vector, it is only necessary to increase the number of unimodal recognition networks and modify the number of initial feature integrations. More importantly, the system proposed in this paper has high security, that is, in real life and applications, it is often difficult for criminals to obtain multimodal information of fingers at the same time. Exploring and developing an attention mechanism more suitable for finger modalities and improving the security of finger multimodal recognition systems are the focus of our research in the next stage of work.

## Acknowledgements

This work is partly supported by National Natural Science Foundation(NNSF) of China under Grant 61873131, and Postgraduate Research & Practice Innovation Program of Jiangsu Province under Grant KYCX20_0828.